\documentclass[twocolumn]{article}
\usepackage{amsmath,amssymb,amsfonts}
\usepackage{algorithmic}
\usepackage{graphicx}
\usepackage{textcomp}
\usepackage{authblk}
\usepackage{sectsty}
\sectionfont{\bfseries\Large\raggedright}
\def\BibTeX{{\rm B\kern-.05em{\sc i\kern-.025em b}\kern-.08em
    T\kern-.1667em\lower.7ex\hbox{E}\kern-.125emX}}
    
\usepackage{biblatex}   
\usepackage{caption}
\usepackage{subcaption}
\usepackage{color,soul}

\newcommand{\eg}{{\it e.g.},~}
\newcommand{\ie}{{\it i.e.},~}
\newcommand{\etal}{{\it et al.}}

\begin{document}

\title{Improving Model's Focus Improves Performance of Deep Learning-Based Synthetic Face Detectors}
\author[1]{Jacob Piland}
\author[2]{Adam Czajka}
\author[3]{Christopher Sweet}
\affil[1]{Department of Computer Science and Engineering, University of Notre Dame, Notre Dame, IN 46556, USA (e-mail: jpiland@nd.edu), Corresponding Author}
\affil[2]{Department of Computer Science and Engineering, University of Notre Dame, Notre Dame, IN 46556, USA (e-mail: aczajka@nd.edu)}
\affil[3]{Center for Research Computing, University of Notre Dame, Notre Dame, IN 46556, USA  (e-mail: csweet1@nd.edu)}
\maketitle

\begin{abstract}
Deep learning-based models generalize better to unknown data samples after being guided ``where to look'' by incorporating human perception into training strategies. We made an observation that the entropy of the model's salience trained in that way is lower when compared to salience entropy computed for models training without human perceptual intelligence. Thus the question: does further increase of model's focus, by lowering the entropy of model's class activation map, help in further increasing the performance? In this paper we propose and evaluate several entropy-based new loss function components controlling the model's focus, covering the full range of the level of such control, from none to its ``aggresive'' minimization. We show, using a problem of synthetic face detection, that improving the model's focus, through lowering entropy, leads to models that perform better in an open-set scenario, in which the test samples are synthesized by unknown generative models. We also show that optimal performance is obtained when the model's loss function blends three aspects: regular classification, low-entropy of the model's focus, and human-guided saliency.
\footnote{This work was supported by the U.S. Department of Defense (Contract No. W52P1J2093009). The views and conclusions contained in this document are those of the authors and should not be interpreted as representing the official policies, either expressed or implied, of the U.S. Department of Defense or the U.S. Government.}
\end{abstract}

\section{Introduction}
\label{sec:introduction}

\begin{figure*}[!t]
\centering
    \includegraphics[width=\textwidth]{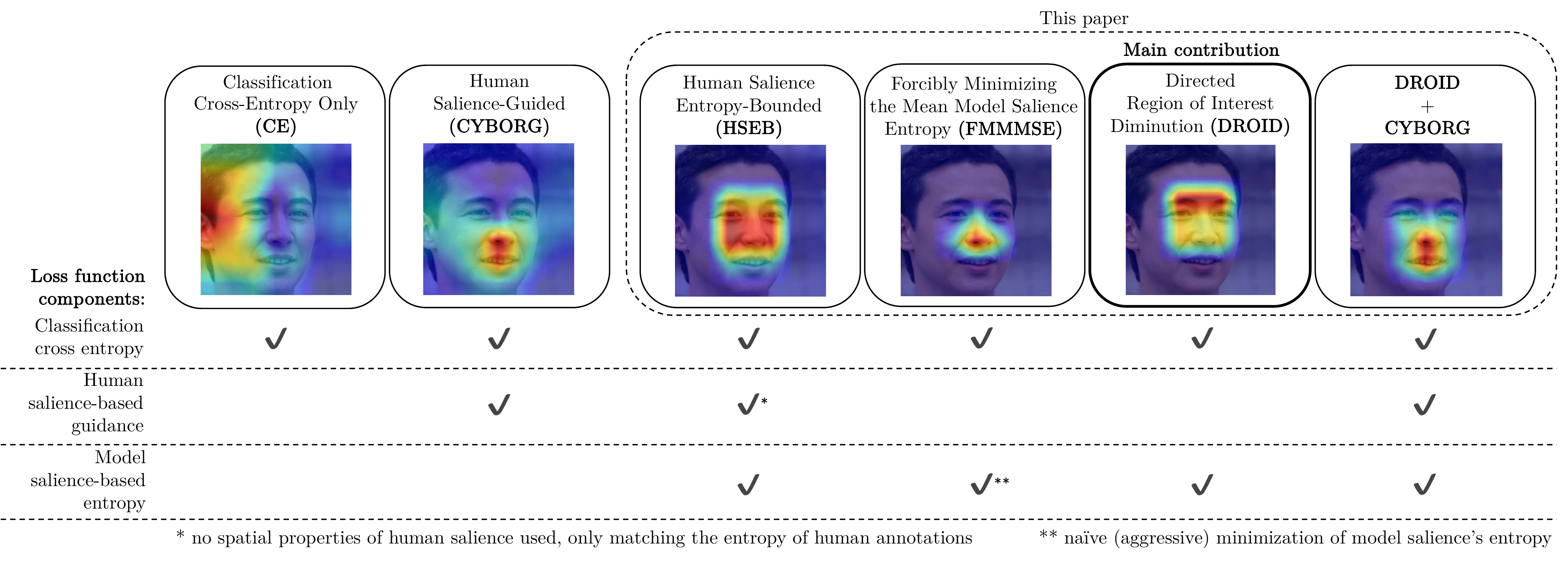}
    \caption{Various approaches to build loss functions for models detecting synthetic samples, and example model's salience (pictures as class activation maps). The most common penalizes the classification cross-entropy ({\bf CE}). Recent approaches suggest penalizing deviations of the model salience from the human salience ({\bf CYBORG} \cite{cyborg2022}). In this paper we investigate how shaping the entropy of the model salience, independently of the human salience, performs in the task of synthetic face detection, and propose {\bf DROID}: the method of directed diminution of model salience's entropy. To get a full picture on possible variants of model and human salience combinations in a single loss function, we explore losses that only match the average entropy of the human annotations and model salience ({\bf HSEB}), and investigate if aggressive minimization of model salience's entropy ({\bf FMMMSE}) may be sufficient to solve the synthetic face detection task. Finally, we combine all three types of losses (classification cross-entropy, human salience-based, and model salience-based ({\bf DROID+CYBORG})}.
    \label{fig:teaser}
\end{figure*}

Modern generative models are able to synthesize non-existing\footnote{With exception, notably when the models ``leak'' identity information found in the training set to the generated samples \cite{Tinsley_WACV_2021, Tinsley_IJCB_2022}} objects, including human faces, with a high (and constantly increasing) degree of realism. Examples of real and synthetic faces, are shown in Fig. \ref{fig:face_examples}.
This makes the fully data-driven discovery of which features are fundamental for automatic synthetic face detection an unfathomable task. 
Human beings, in turn, are naturally adept at finding and recognizing faces, and exceptionally sensitive to minuscule aberrations in face appearance. 
Thus, using human salience to guide the process of training of deep learning-based synthetic face detectors has proved to increase the generalization of such models (to unknown data) \cite{cyborg2022}. 
This is achieved by focusing the model on features identified by humans as being prominent, instead of on features accidentally correlated with class labels. 

An interesting observation we made about human salience-trained models is that the average entropy of a model's salience (estimated by Class Activation Map [CAM] \cite{Zhou_CVPR_2016}) is lower than entropy of salience of models trained by a standard minimization of classification cross-entropy loss. Hence, an immediate question: how does the entropy of a model's salience (regulated through \eg a loss function) relate to the model's generalization, and -- as a consequence -- to its strength of detecting synthetic face images? 
This paper answers this question by exploring several variants of shaping the entropy of the model's salience with and without human guidance embedded into training. 
We show that appropriate entropy of the model's salience (not ``too large'' to keep the model's focus, and not ``too small'' to prevent over-fitting to specific features) allows to build an effective synthetic face detector, generalizing to samples generated by unknown Generative Adversarial Networks (GAN) much better than state of the art solutions.
In particular, we explore the loss functions incorporating the entropy with different mixtures of both human saliency \cite{cyborg2022} and model's saliency.
% We combine the loss functions of the appropriate middle-of-the-road low entropy model with the CYBORG component of the loss function from \cite{cyborg2022}.
% In this way we explore the naive combination of low entropy focus and human saliency guided learning.

% ***AC: This is good. I would move it to the section where this is done, and mention the off-center experiments in contributions below
% Finally, if we view reducing the entropy of the model's saliency as increasing the model's focus, it naturally leads us to ask if the model is becoming too focused. 
% The facial data we use to explore our low-entropy variants is from an already established dataset \cite{cyborg2022} with centered, closely framed faces. 
% We test the over-focusing of the models with a new dataset of off-center faces.

We define the following {\bf research questions} (RQ), and organize the paper in a way to answer these questions. All questions relate to models trained to detect synthetically generated faces by a GAN model held out for testing (hence, unknown during training). When we speak about human salience, we assume it's available in a form of regions that humans annotated, as in previous experiments by Boyd \etal, who introduced the human perception-based guidance into the loss function called CYBORG~\cite{cyborg2022}.

\begin{enumerate}
    \item[{\bf RQ1:}] Let's assume that human saliency information is not available, but we can estimate the average entropy of human salience. Does requesting the average entropy of the model's and human's saliencies to match increase the performance, compared to a performance of a model trained traditionally with just cross-entropy loss? (see {\bf HSEB} variant in Fig. \ref{fig:teaser})

    \item[{\bf RQ2:}] If the answer to RQ1 is affirmative, what if we aggressively request the minimum possible entropy of the model's saliency? The intuition behind this approach is that there are perhaps single or well-localized and strong-enough features that are sufficient to solve the synthetic face detection task. (see {\bf FMMMSE} variant in Fig. \ref{fig:teaser})

    \item[{\bf RQ3:}] Is there a better strategy to control the entropy of the model's saliency than requesting a specific entropy (as in RQ1) or aggressively minimizing such entropy (as in RQ2)? (see {\bf DROID} variant in Fig. \ref{fig:teaser})

    \item[{\bf RQ4:}] Finally, does the model benefit from combining the human saliency-guided training with non-human-guided control of the model salience's entropy? (see {\bf DROID+CYBORG} variant in Fig. \ref{fig:teaser})

\end{enumerate}

\section{Related Work}
%List of things to reference:
%\begin{itemize}
%    \item changing the loss function on a framework can improve performance; it is a valid experiment to just try new loss functions \cite{loss_function_swap2021}
%    \item the work of cyborg and some of its related works \cite{cyborg2022} and 
%    \item the paper that defines entropy cam \cite{cam_entropy2022}
%\end{itemize}

Image manipulation and the creation of fake images poses a serious threat in terms of security \cite{50wang2020cnn,4botha2020fake,7chesney2019deepfakes}. 
A particularly well known and socially relevant example is that of creating facial data \cite{faceswap}. 
The wholesale generation of synthetic faces using GANs was first demonstrated in 2014 \cite{16goodfellow2014generative} and there have been numerous examples since \cite{25karras2017progressive,stylegan_ada2020,stylegan2019,stylegan2020,stylegan2021}. 

There have been many efforts to detect fake faces \cite{46kitware,24Gandetector,52ganscanner,17gandetection,33mandelli2020training,50wang2020cnn} and frequency domain analysis has had success in detecting synthetic faces \cite{41qian2020thinking,18grieggs2021measuring}.
However, deep neural networks can achieve a recall of synthetic faces of 99\% due to the effectively endless supply of generated fake face samples \cite{48tariq2019gan} even if the overall accuracy is not perfect. One of these deep neural networks is DenseNet \cite{densenet121_2017}, which we use as the backbone (pre-trained framework and starting point) for the models tested in this paper.

However, there remains a level of inexplanability to these deep learning methods that can be alleviated by comparison to observed, human experts detecting synthetic faces \cite{42richardwebster2018visual}.
Furthermore, while machine learning accuracy is essentially always at least as good as human accuracy \cite{37o2012comparing}, human pyscophysics has aided in deep learning tasks such as handwriting \cite{18grieggs2021measuring}, natural language processing \cite{53zhang2020human}, and scene description \cite{20he2019human,23huang2021specific}.
Specifically in biometrics (including synthetic face detection), human saliency has been shown to compliment machine saliency \cite{49trokielewicz2019perception,11czajka2019domain,35moreira2019performance,cyborg2022}. 
Of particular importance to this study is the CYBORG study \cite{cyborg2022} which is covered in detail in Sec. \ref{sect:CYBORG_entropy}.

Finally, the idea of measuring CAM entropy as a meaningful way to improve model explainability is a relatively new idea, but established in \cite{cam_entropy2022}.
\section{Human saliency-guided training reduces entropy of model's salience}
\label{sect:CYBORG_entropy}

\begin{figure}
\centering
     \begin{subfigure}[b]{0.11\textwidth}
     \centering
         \includegraphics[width=\textwidth]{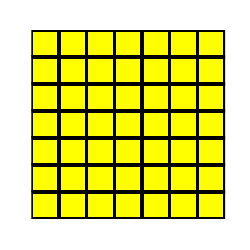}
         \caption{3.892}
         \label{fig:ex1}
     \end{subfigure}
     \hfill
     \begin{subfigure}[b]{0.11\textwidth}
     \centering
         \includegraphics[width=\textwidth]{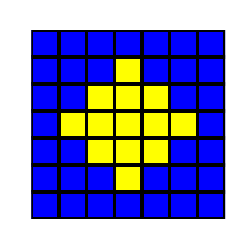}
         \caption{2.566}
         \label{fig:ex2}
     \end{subfigure}
     \hfill
     \begin{subfigure}[b]{0.11\textwidth}
     \centering
         \includegraphics[width=\textwidth]{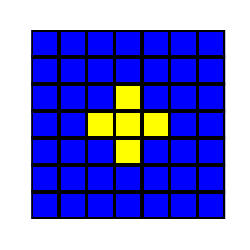}
         \caption{1.613}
         \label{fig:ex3}
     \end{subfigure}
     \hfill
     \begin{subfigure}[b]{0.11\textwidth}
     \centering
         \includegraphics[width=\textwidth]{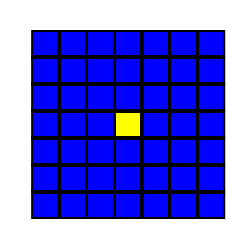}
         \caption{0.018}
         \label{fig:ex4}
     \end{subfigure}
    \caption{{\bf (Entropy of the probability density map)} Example CAMs for a 7-by-7 grid that have been normalized to sum to 1. Underneath each CAM is its corresponding entropy value. In each CAM focus is mapped to a yellow-to-blue color scale with yellow corresponding to its highest value (i.e., focus) and blue to its lowest. This demonstrates how Shannon entropy shrinks as the number of pixels focused on shrinks.}
    \label{fig:entropy_examples}
\end{figure}

The human perception-guided training aims at minimizing the distance between the model's saliency maps and their respective human saliency maps. % as shown in the CYBORG loss function. 
For instance, in the example CYBORG approach \cite{cyborg2022}, the loss function is composed of two terms: the human perception loss component (the Mean Squared Error between the human salience and the model's salience), and the classification loss component (regular cross-entropy):
\begin{equation}
\label{eqn:CYBORGloss}
\begin{split}
\mathcal{L}_{CYBORG}&=\frac{1}{K} \sum_{k=1}^{K} \sum_{c=1}^{C}\mathbf{1}_{yk \in C_c}\\
&\Big[(1-\alpha)\underbrace{MSE(\mathbf{s}_k^{(h)},\mathbf{s}_k^{(m)})}_{\text{human perception component}}\\
&-\alpha \underbrace{\log p^{(m)}(y_k \in C_c}_{\text{classification component}})\Big]
\end{split}
\end{equation}
where $K$ is the number of samples in a batch, $C$ is the number of classes, $y_k$ is a class label for the $k$-th sample, $\mathbf{1}$ is a class indicator function equal to 1 when $y_k \in C_c$ and 0 otherwise, $\mathbf{s}_k^{(h)}$ is the human saliency map, $\mathbf{s}_k^{(m)}$ is the model's saliency map calculated for the $k$-th sample, and $\alpha$ is the parameter weighting human-based and cross-entropy-based loss components. Boyd \etal~\cite{cyborg2022} selected Class Activation Mechanism (CAM) \cite{Zhou_CVPR_2016} to approximate model's salience $\mathbf{s}^{(m)}$, and normalized both the human and model saliency maps to [0,1]. We follow the same strategy in this work.
% 
% In \cite{cam_entropy2022} the authors consider the CAM as a probability density map and calculate Shannon's Entropy \cite{shannon_entropy} to get a measure of the magnitude of the information it contains. 
% ***AC: something is wrong here -- We can further interpret the CAM entropy as the number of pixels of equally high probability, \ie the number of pixels the model is focusing on. For example an entropy of 0 would have a single pixel of probability 1 (focusing on one pixel). A maximum entropy value indicates an equal probability for all pixels (focusing equally on all pixels).
% 
% Since our intuition is that the human saliency maps focus on specific areas we might expect that they have low Shannon entropy and this is borne out by our experiments as seen in Figure \ref{fig:densenet_entropy}. 

The entropy $H$ of the CAM (or salience map) $\mathbf{s}$ is:
\begin{align}
\label{eqn:CAMentropy}
H = \sum_{i=1}^{h} \sum_{j=1}^{w} -\mathbf{s}(i,j)\log \mathbf{s}(i,j),
\end{align}
where $h$, $w$ are height and widht of the salience map $\mathbf{s}$, respectively, and $\mathbf{s}$ is normalized to formally express the probability distribution related to the concept of ``focus'':
\begin{align}
\sum_{i=1}^{h} \sum_{j=1}^{w} \mathbf{s}(i,j)=1, \qquad 0\le \mathbf{s}(i,j) \le 1 \qquad \forall i,j.
\end{align}

Fig. \ref{fig:entropy_examples} illustrates a few example $7\times 7$ salience maps $\mathbf{s}$ and their corresponding entropy. For instance, an entropy of 3.89 would correspond to all 49 locations of equal probability of $\approx 0.02$ (\ie the model is focusing on all pixels equally, Fig. \ref{fig:entropy_examples}(a)). An entropy of $\approx 0.02$ would correspond to a single location of probability 1.0.  

\begin{figure}[!htb]
\centering
    \includegraphics[width=0.9\linewidth]{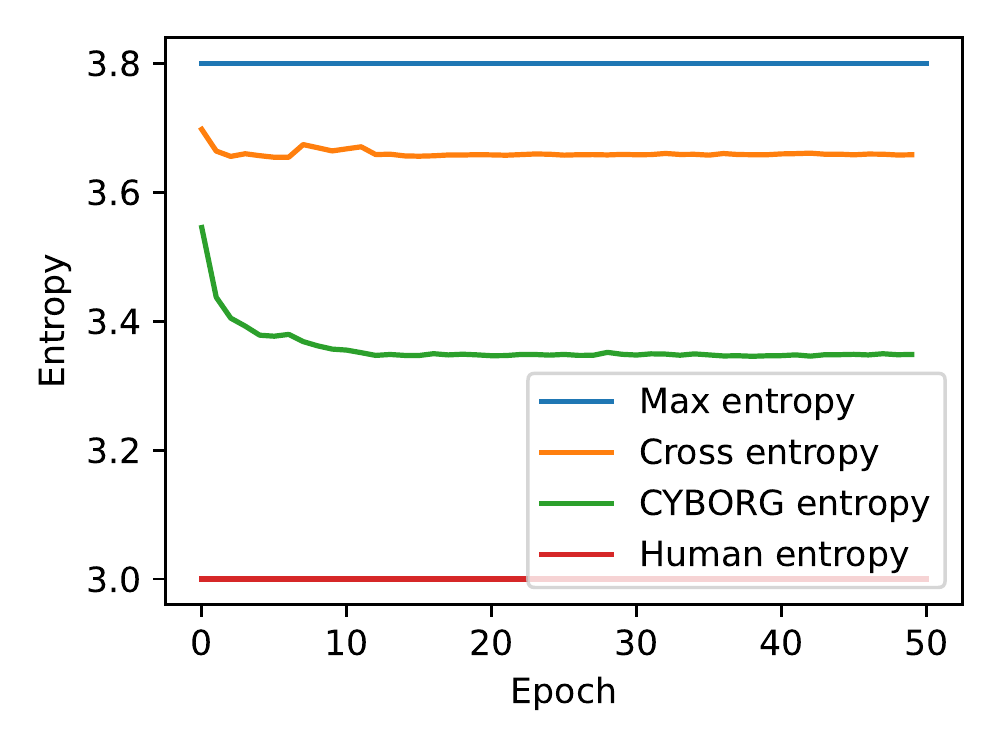}
    \caption{Comparison of entropy of the model's salience $\mathbf{s}$ (estimated via Class Activation Map) over training epochs for a DenseNet model trained using classical cross-entropy and human-guided (CYBORG) approaches. Maximum entropy for a $7\times7$-element salience and the human entropy (measured in synthetic face recognition tasks) are also shown for reference. It can be seen that human guidance during training (CYBORG) ends up with lower model's salience entropy, compared to cross-entropy.}
    \label{fig:densenet_entropy}
\end{figure}

Fig. \ref{fig:entropy_examples} is certainly a toy example, and it is more interesting to observe how the entropy $H$ of actual salience $\mathbf{s}$ estimated for models trained in various ways changes during training. We trained DenseNet \cite{densenet121_2017} with both regular cross-entropy loss, and CYBORG human saliency-guided loss, and compared the entropy of the resulting model salience maps with entropy of human salience (computed directly from human annotations). As we see in Figure \ref{fig:densenet_entropy}, an average salience entropy for a model trained with cross-entropy loss is around 3.65. That corresponds to approx. 38-element (out of 49 for a $7\times7$ Class Activation Map) focus area. For model trained with CYBORG loss, however, the model's salience entropy goes down to 3.30 (what corresponds to a 27-element focus). For comparison, human-annotated salience maps have an average entropy of 3.0. Looking at the saliency maps in Fig. \ref{fig:teaser} we see a reasonable correlation with these figures. This experiment, serving as a segue to Sec. \ref{sect:low_entropy_methods}, suggests that (a) human-guided training decreases the model's salience entropy, and (b) there is a negative correlation between the salience entropy and the model's performance.

% For our experiments using Densenet \cite{densenet121_2017} we have CAM of dimension 7 x 7 which has a maximum entropy value of 3.89 (all pixels of equal probability $p$, $\approx p=0.02$).
\section{Proposed low entropy models}
\label{sect:low_entropy_methods}

Section \ref{sect:CYBORG_entropy} demonstrated that Shannon's entropy of the model's salience is reduced when the network is guided towards important features during training. % CAMs gave a measure of how focused the network was on areas within the images. 
% From our experiments we believe that Shannon's entropy for CAM provides 
% This provides an insight into how ``focused'' the model is. 
Extending this insight, we propose to examine several methods of minimizing entropy of the class activation maps (serving as an estimator of the model salience) and analyze which methods increase the generalization capabilities of the model in the task of synthetic face detection. An important note is that the proposed approaches are not limited to synthetic face detection task, and can be applied to problems, in which human perceptual capabilities may be utilized in model's training.

The proposed overall approach can be seen as a generalization of the human-guided CYBORG training introduced by Boyd \etal \cite{cyborg2022}. We do this by replacing the human perception component in Eq. (\ref{eqn:CYBORGloss}) with a more generic {\it salience entropy control component} $\mathcal{L}^{(sec)}$, namely:

\begin{equation}
\begin{split}
\mathcal{L}&=\frac{1}{K} \sum_{k=1}^{K} \sum_{c=1}^{C}\mathbf{1}_{yk \in C_c}\\
&\Big[(1-\alpha)\mathcal{L}^{(sec)}_k-\alpha \underbrace{\log p^{(m)}(y_k \in C_c}_{\text{classification component}})\Big]
\end{split}
\label{eqn:low_entropy_loss}
\end{equation}
%
%The resulting loss function follows the format in Eq. \ref{eqn:low_entropy_loss}.
%\begin{equation}
%\label{eqn:low_entropy_loss}
%\begin{split}
%\mathcal{L}_{CC} &=\frac{1}{K} \sum_{k=1}^{K} \sum_{c=1}^{C} \mathbf{1}_{yk \in C_c}\\
%&\qquad[(1-\alpha)(\text{Entropy loss component})\\
%&\qquad-\alpha(\text{classification component})]\\
%\end{split}
%\end{equation}
%
where all variables have the same meaning as in Eq. (\ref{eqn:CYBORGloss}). % and the three different low entropy loss components are described separately in Eq. \ref{eqn:HBloss}, Eq. \ref{eqn:FMMMSEloss}, and Eq. \ref{eqn:DROIDloss}.
Further in this Section we investigate three different approach to building $\mathcal{L}^{(sec)}$

The first approach, Human Salience Entropy Bound ({\bf HSEB}), which is directly related to research question {\bf RQ1} aims at matching the Shannon's entropy of the model's and human's salience maps:

\begin{equation}
\mathcal{L}^{(sec)}_k = MSE(H_k^{(m)}, H_k^{(h)})
\end{equation}
where $H_i^{(m)}$ and $H_i^{(h)}$ are the entropies of the CAMs and human saliency maps, as defined in Eq. \ref{eqn:CAMentropy}, respectively, and averaged over all samples within the $i$-th batch. Note that in this approach we do not guide the model ``where to look'' and only request the model to achieve a similar salience's entropy as observed for humans who annotated the same training samples. The motivation for this is to give the model more flexibility in choosing salient features and exploring an approach in which the exact human saliency maps are not available, but instead we know the estimated value of their entropy. If the generalization capabilities of such approach is competitive, we could potentially replace the need of collecting human salience maps with an estimated scalar entropy value of such maps.

% The results for entropy can be seen in Fig. \ref{fig:densenet_entropy} as HSEB where the CAM entropy indeed asymptotically approaches that of the human saliency entropy. 
% Results discussed in Sec. \ref{sec:results} show an improvement in performance over cross-entropy and CYBORG.

The observed results obtained for the HSEB approach (discussed in details in Sec. \ref{sec:results}) suggest that this way of limiting the model salience's entropy allows to further improve the performance. Following this, in some sense na\"ive approach, we explored the way to aggressively minimize the model's salience, called Forcibly Minimizing the Mean Model Salience Entropy ({\bf FMMMSE}), directly addressing the research question {\bf RQ2}:

\begin{equation}
\mathcal{L}^{(sec)}_k = H_k^{(m)}
\end{equation}
with $H_k^{(m)}$ defined in Eq. \ref{eqn:CAMentropy}. This method, as we will see later, not surprisingly overfits to the training data, suggesting that the model salience's entropy cannot be minimized too aggressively as it promotes searching for spurious features correlated with the class category (that is, what we want to avoid. This takes us to the last, and the most effective approach to select ${L}^{(sec)}$, which investigates a middle ground between being close to human entropy and minimum entropy of model's salience: Directed Region Of Interest Diminution ({\bf DROID}). DROID minimizes the log CAM entropy, which is similar to the FMMMSE approach, but with less of a penalty on higher entropy to avoid over-focusing:

\begin{equation}
\mathcal{L}^{(sec)}_k = \log\big(H_k^{(m)}\big)
\label{eqn:droid}
\end{equation}
for each sample $k$ in a batch.
\section{Proposed combination of human-saliency and low entropy}
\label{sec:combination}

%\noindent
%{\bf *** Adam is here ***}
%\vskip1mm

% 
% 
In Section \ref{sect:low_entropy_methods} we saw that low entropy models force a model's focus on few features.
However, there is no guarantee that these will be the most important, or even useful features.
CYBORG approach, as we saw in Sec. \ref{sect:CYBORG_entropy}, uses human saliency to guide a model to important features, but the features to focus on are desired to be matched with those with humans, including their number. We hypothesise that if put together, human saliency should guide the model to important features while low entropy should force the model to focus on only the most important features.
We thus propose a combination of these two approaches (called {\bf CYBORG+DROID}) as an exploratory test, by using DROID as the low entropy component and CYBORG as the human-saliency component, addressing research question {\bf RQ4}:

\begin{equation}
\label{eqn:CYBORGDROIDloss}
\begin{split}
\mathcal{L}_{CYBORG+DROID}=\frac{1}{K} \sum_{k=1}^{K} \sum_{c=1}^{C} \mathbf{1}_{yk \in C_c} & \\
\Big[\alpha\underbrace{MSE(\mathbf{s}_k^{(h)},\mathbf{s}_k^{(m)})}_{\text{human perception component}}
+\beta\underbrace{\log\big(H_k^{(m)}\big)}_{\text{low entropy component}}&\\
-\gamma \underbrace{\log p_{model}(y_k \in C_c}_{\text{classification component}}\Big]
\end{split}
\end{equation}
for each sample $k$ in a batch of size $K$.

% because it will be shown that it is the highest performing low entropy model that does not over focus (Sec. \ref{sec:centered_results} and Sec. \ref{sec:offset_results} respectively).

As we see, CYBORG+DROID follows a similar loss function format as the other low entropy models, excepting that there is both a salience entropy
control component (specifically, DROID, defined by \eqref{eqn:droid}) and a CYBORG loss component (defined by \eqref{eqn:CYBORGloss}), each with their own weights. 

In the original CYBORG study it was determined that CYBORG is largely unaffected by changes to the coefficients in front of the loss components ($\alpha$ and $\gamma$ in \eqref{eqn:CYBORGDROIDloss}). 
In this study, we did explore using various weight values for the components of our proposed low entropy models, and they too were largely unaffected. However, the combination model, CYBORG+DROID, is affected by changes in component weights in terms of final average CAM entropy, performance score, and over focusing. We found that $\alpha=0.5$, $\beta=0.3$, and $\gamma=0.5$ in Eq. \ref{eqn:CYBORGDROIDloss} are optimal for the CYBORG+DROID approach.

% \begin{figure}[ht]
% \centering
%     {\includegraphics[width=6.8cm]{figures/human_cam_comp_centered.png} }
% \caption{Comparison between CAM and human saliency maps illustrating the correlation between high probability pixels and that predicted by entropy. Each row is three, random, human annotated images from the CYBORG training set. The y-axis indicates the model and whether the row contains human annotations or CAMs.}
% \label{fig:DROID_cams}
% \end{figure}

% In Figure \ref{fig:teaser} we see an example of the CAMs generated from the different loss functions overlayed on their respective sample images.

% \begin{figure}[ht]
% \centering
%     {\includegraphics[width=6.8cm]{figures/samples_per_model.png} }
% \caption{Illustration of the CAM focus overlayed onto a sample of images for each of our tested models. The x-axis specifies the model. For each row, the left column is a real sample (from CelebA-HQ) and the right column is a fake sample (from stylegan 3). Yellow, green, and purple represent decreasing focus in that order.}
% \label{fig:DROID_overlay}
% \end{figure}

\label{sec:datasets}

\begin{figure*}
\centering
     \begin{subfigure}{\textwidth}
     \centering
         \includegraphics[width=5.1in]{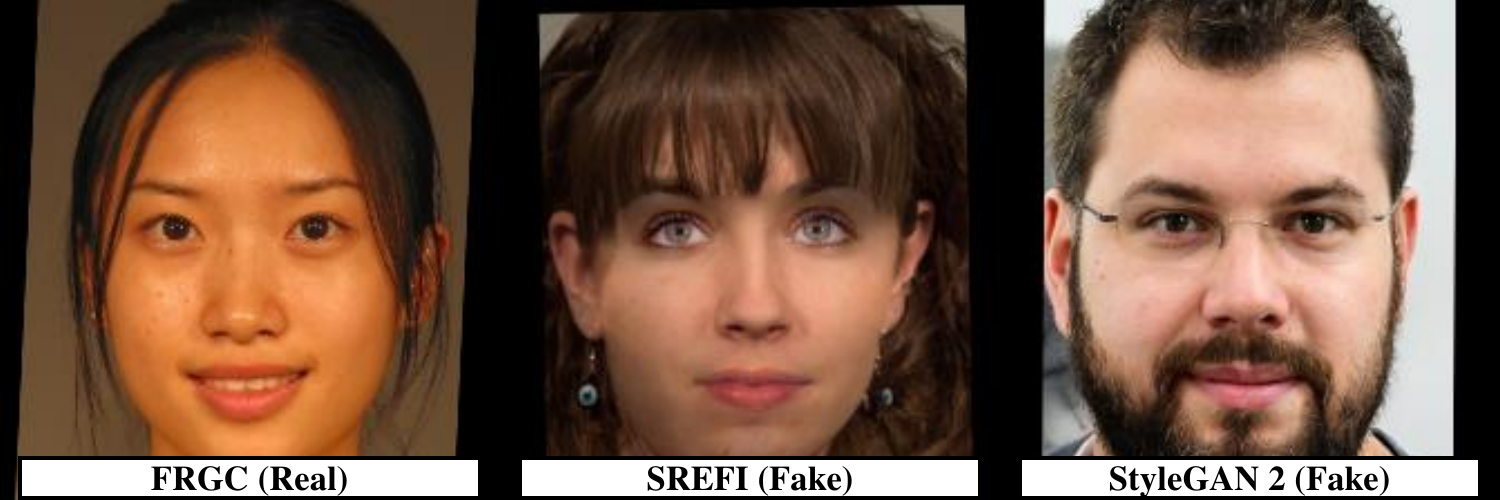}
         \vskip-1mm
         \caption{Training set samples}
         \label{fig:trainset}
     \end{subfigure}
     \vskip3mm
     \begin{subfigure}{\textwidth}
     \centering
         \includegraphics[width=5.1in]{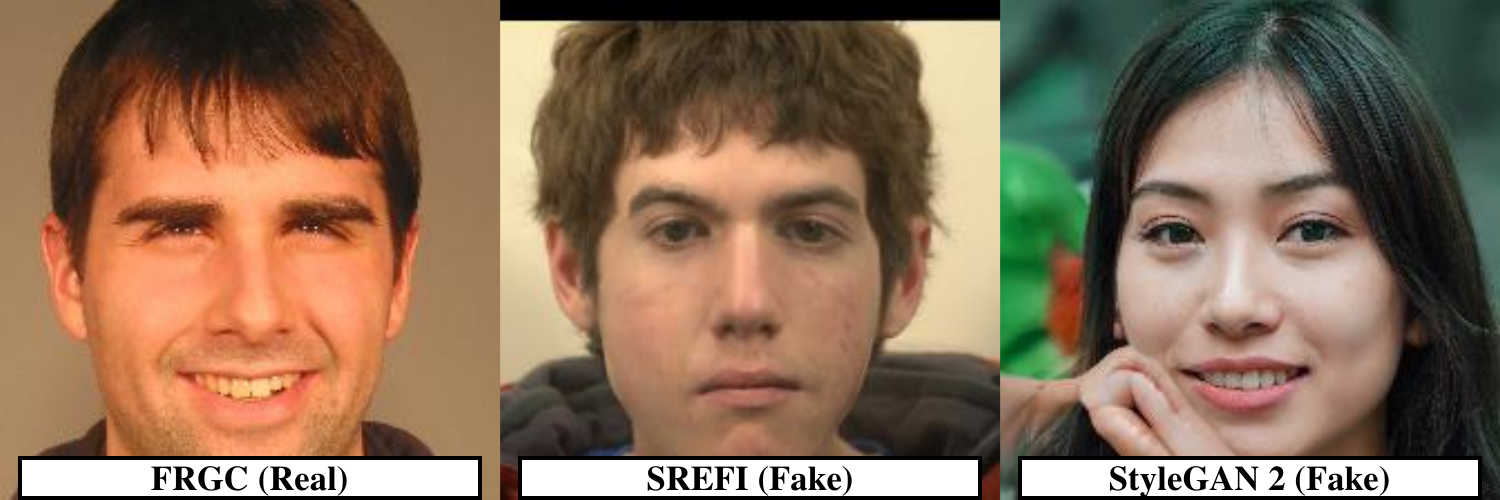}
         \vskip-1mm
         \caption{Validation set samples}
         \label{fig:valset}
     \end{subfigure}
     \vskip3mm
     \begin{subfigure}{\textwidth}
     \centering
        \includegraphics[width=6.8in]{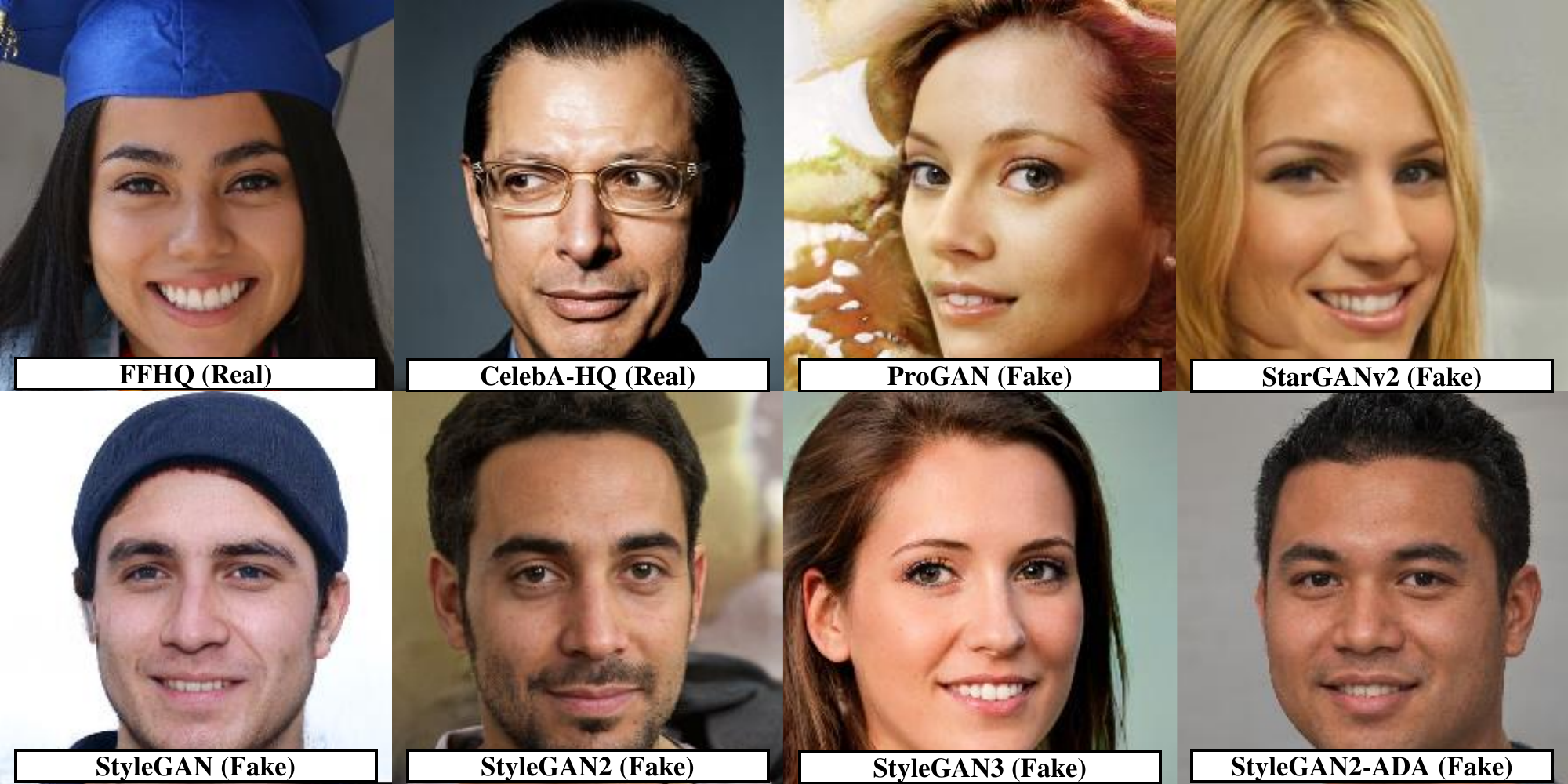}
        \vskip-1mm
        \caption{Test set samples}
        \label{fig:testset}
     \end{subfigure}
    \caption{Examples of real and fake faces from each of the datasets or generators used in the a) training set, b) validation set, and c) testing set.}
    \label{fig:face_examples}
\end{figure*}

\section{Experimental Setup}
\subsection{Experiment descriptions}
% Here we generally discribe the experimental procedures.
% Specific parameters are listed in Sec. \ref{sec:exp_params}.

We conduct four experiments, one to address each of the four research questions: explicitly requesting the model and human saliency match in terms of CAM entropy value (HSEB approach, addressing \textbf{RQ1}), aggressively requesting the minimum CAM entropy possible (FMMMSE approach, addressing \textbf{RQ2}), less aggressively requesting the minimum CAM entropy with a log-loss function (DROID approach, addressing \textbf{RQ3}), and the combination of low model's entropy request and human saliency-based guidance (DROID + CYBORG, addressing \textbf{RQ4}). %, and \textbf{RQ5} testing the low entropy models for over-focusing.  

The same experimental format is used in all four experiments.
% The purpose of 
In each experiment we compare the performance of one of the low entropy models, or the CYBORG+DROID model, to baseline cross-entropy and state-of-the-art CYBORG in the task of synthetic face detection. While synthetic face detection was chosen as an example domain, all the considerations remain valid for other visual tasks in which humans are competent. An increase in performance from the baseline or state-of-the-art compared to the low entropy model, where only the loss functions distinguish between the models, will indicate that training a model with a constraint put on its CAM entropy is beneficial. The performance of each model is measured using area under the Receiver Operating Characteristic (AUROC) based on sigmoid scoring. In all cases, cross-entropy-based models serve as our baseline for comparisons, and CYBORG models as the state-of-the-art for comparisons.

% Is the below paragraph repeating what was said above?
% A number of model samples are trained using the HSEB, FMMMSE, DROID, and CYBORG+DROID loss functions seperately. 
% We also train models for cross-entropy and CYBORG and we use the same cross-entropy and CYBORG models for comparison in each of the \textbf{RQ1-4} experiments.

\subsection{Experiment parameters}
\label{sec:exp_params}

For training, we follow the experimental procedure established in \cite{cyborg2022}, excepting learning rate and number of epochs.
Both of these changes were done to more thoroughly explore the behavior of our low entropy models.
Specifically, all models are trained with a constant learning rate of 0.002 for a period of 150 epochs using Stochastic Gradient Descent and the weights chosen for the final model are those offering the highest validation accuracy.
All samples are instantiated from the DenseNet-121 model pre-trained on ImageNet dataset \cite{densenet121_2017}.
The training and validation sets are constant for all models as described in Sec. \ref{sec:datasets}.
% In order to generate error statistics, we trained ten samples for each of the models considered in this paper, including the baseline cross-entropy and state-of-the-art CYBORG. 
To assess the uncertainty associated with randomness of the training process, we train ten instances of each model with the same training data but with different seeds, and use an average AUROC with standard deviation margins in comparisons.

The weighting for loss components is as follows.
For the cross-entropy baseline, classification loss is $\alpha=1.0$.
For CYBORG, HSEB, FMMMSE, and DROID, the weighting for all loss components is equal: $alpha=0.5$.
For CYBORG+DROID, the weighting for the classification loss is $\gamma=0.5$, the low entropy DROID component is $\beta=0.3$, and the CYBORG human component is $\alpha=0.5$.
%
% For testing, in the experiments addressing \textbf{RQ1-4} we use the testing dataset used in \cite{cyborg2022}.
%For the experiment addressing \textbf{RQ5}, we use a new dataset described in Sec. \ref{sec:offset_dataset}. 
% Our metric for comparing model performance is AUROC with sigmoid scoring.
% For our Bayesian estimation we use the defaults of the R \cite{R}, BEST package \cite{BESTmcmc}.

\subsection{Datasets}

For training each model we use the established face image datasets, split into disjoint training, validation, and testing datasets, in the same way as proposed in \cite{cyborg2022}. 
Figure \ref{fig:face_examples} shows synthetic and real face image examples from each dataset.

The {\bf training set} consists of 1821 training samples (919 real and 902 synthetic). 
Real samples originate from the Face Recognition Grand Challenge (FRGC) dataset, and synthetic samples are generated for this dataset using the “synthesis of realistic face images" (SREFI) method \cite{SREFI} and StyleGAN2 \cite{stylegan2020}.

The {\bf validation set} consists of 20,000 samples (10,000 real and 10,000 synthetic). 
As with the training set, real validation samples are taken from FRGC, and synthetic validation samples are generated with SREFI and StyleGAN2.
Note that separate images were generated from SREFI and StyleGAN2 for the training and validation datasets.

Finally, the {\bf test set}, kept the same for all experiments, consists of 700,000 samples. There are 600,000 synthetic samples, 100,000 from each of the following GANs: ProGAN \cite{progan2019}, StarGANv2 \cite{stargan2018}, StyleGAN \cite{stylegan2019}, StyleGAN2 \cite{stylegan2020}, StyleGAN3 \cite{stylegan2021}, and StyleGAN2-ADA \cite{stylegan_ada2020}. 
The images from three of these sources were pre-generated: the ProGAN images are from \cite{27progan_source} and then the StyleGAN and StyleGAN2 images are from their GitHUB repositories. 
Samples from the remaining three sources, StyleGAN3, StyleGAN2-ADA, and StarGAN were generated for this dataset \cite{cyborg2022}.
There are 100,000 real samples: 30,000 images from CelebA-HQ \cite{celeba_hq2017}, and 70,000 images from Flicker-Faces-HQ (FFHQ) \cite{FFHQ2018}.

\section{Results}
\label{sec:results}
\subsection{Model Entropy vs Performance}

\label{sec:final_entropy_values}
\begin{figure*}[!h]
    \centering
    \includegraphics[width=\textwidth]{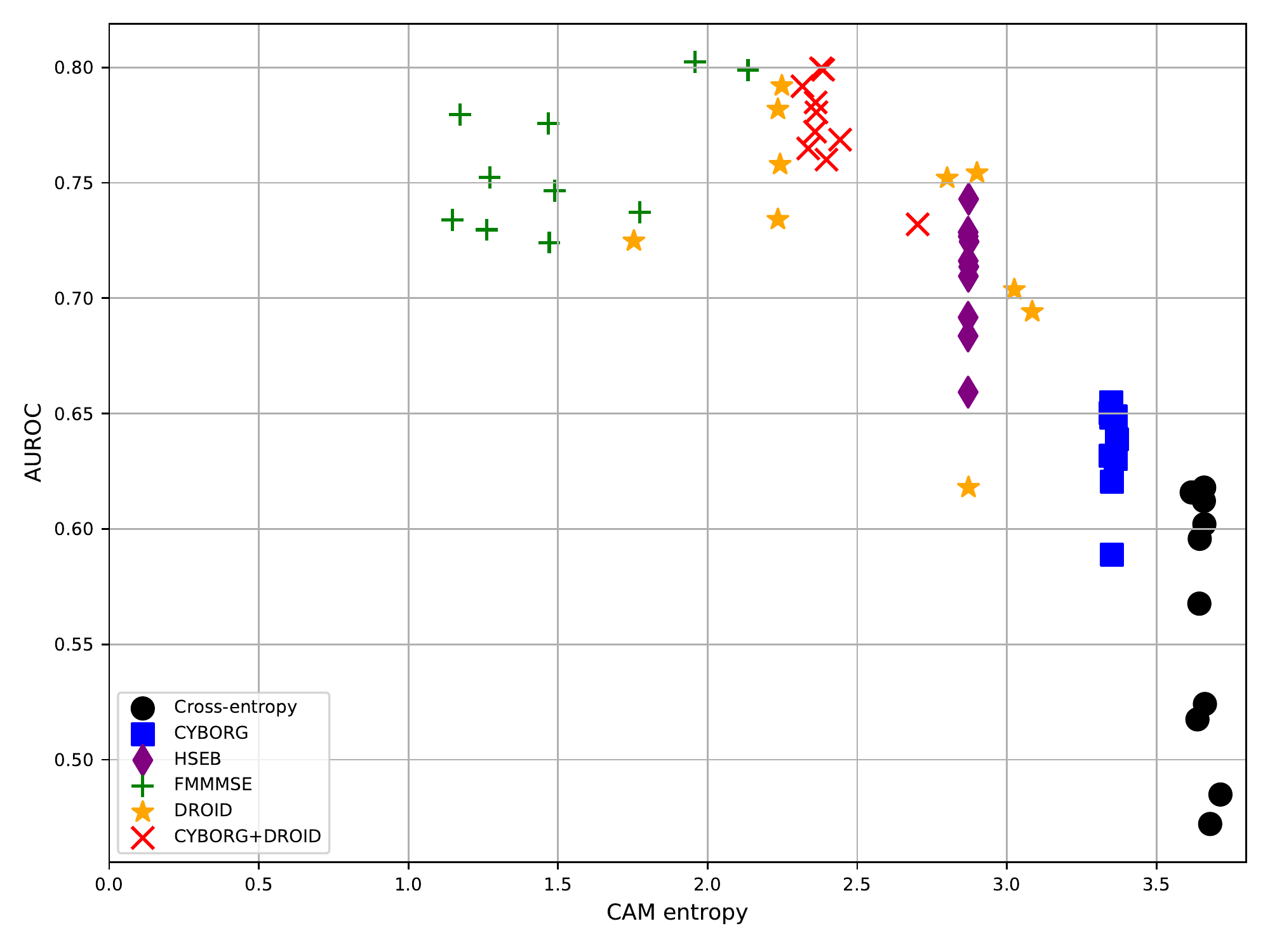}
    \caption{The CAM entropy and AUROC performance for each of the 10 samples of each of the six models tested. The y-axis is AUROC for each sample using sigmoid scoring and the x-axis is the CAM entropy for that sample. Note that generally the CAM entropy decreases for each model in the following order: cross-entropy, CYBORG, HSEB, DROID/CYBORG+DROID, and FMMMSE. The highest AUROC performances are not at either entropy extreme, but close to the middle range of 2.0 to 2.5.}
    \label{fig:entropy_by_model}
\end{figure*}

Motivated by \cite{cam_entropy2022}, we observed in Section \ref{sect:CYBORG_entropy} that the CAM entropy for models trained with Cyborg was lower than those trained with Cross-entropy. This is unsurprising since Cyborg tries to match the CAMs to human salience maps, which themselves have lower entropy (as a probability density map) as seen in Figure \ref{fig:densenet_entropy}. 

Since our proposed methods, HSEB, FMMMSE and DROID are designed to lower CAM entropy we want to determine if there is a more general correlation between the AUROC scores and CAM entropy. In addition, from Figure \ref{fig:entropy_examples}, CAM entropy indicates how focused the CAM is and it seems unlikely that very small values of CAM entropy would be favorable, so there may be an optimal value for a given model and dataset.

Experimentally we compare the CAM entropy/AUROC performance of the baseline cross-entropy and  state-of-the-art CYBORG with HSEB, FMMMSE, DROID and a combined method, CYBORG+DROID.
% We start our analysis with comparing all six variants (baseline cross-entropy, state-of-the-art CYBORG, HSEB, FMMMSE, DROID and combination CYBORG+DROID) considering jointly their performance (AUROC) and the entropy of model's salience (CAM entropy). 
For each model  
% We do not neccesarily use the weights from the last epoch during training for testing. Instead, 
we use the weights from the epoch with the highest validation accuracy during training, and then do all analyses described later on the sequestered test subset. %, so we must examine the CAM entropy of the models used in testing to ensure that our low entropy models are actually lowering entropy.

Fig. \ref{fig:entropy_by_model} shows the results for ten training runs for each model type. We observe good correlation between lower CAM entropy and increasing AUROC for CAM entropy values above 2.0 with a ``Sweet Spot'' between 2.0 and 2.5. The only method that can achieve CAM entropy values lower than 2 is FMMMSE but AUROC is generally lower, giving credence to the idea that there may be a point where too focused a CAM may detract from the AUROC scores. For the various methods, their highest singular AUROC score increases in the following order: baseline cross-entropy, state-of-the-art CYBORG, HSEB, DROID, CYBORG+DROID, and FMMMSE. We note that although FMMMSE achieved the highest AUROC score, it did so with a moderate CAM entropy value around 2.0, its other results were not as good as CAM entropy was forced lower.

Summarizing, we can make a number of important conclusions based on Fig. \ref{fig:entropy_by_model}. First, reducing CAM entropy, even without the guidance of human salience, gives significant improvement in AUROC scores. Second, training approaches that aggressively minimizing a model's CAM entropy (like FMMMSE) do not end up with models offering the best performance capabilities, and their performance varies greatly across different training runs. 
% This is a consequence of forcing the model to focus on small areas, without the guidance (other than minimization of classification loss) which features are salient. 
Finally, adding complementary human guidance, as in DROID+CYBORG approach, stabilizes the model in terms of CAM entropy across training runs, and offers the best performance capabilities, translating to the highest average AUROC for DROID+CYBORG in Fig. \ref{fig:entropy_by_model}.

\subsection{Addressing research questions}
\label{sec:centered_results}

\begin{figure}
    \centering
    \includegraphics[width=.48\textwidth]{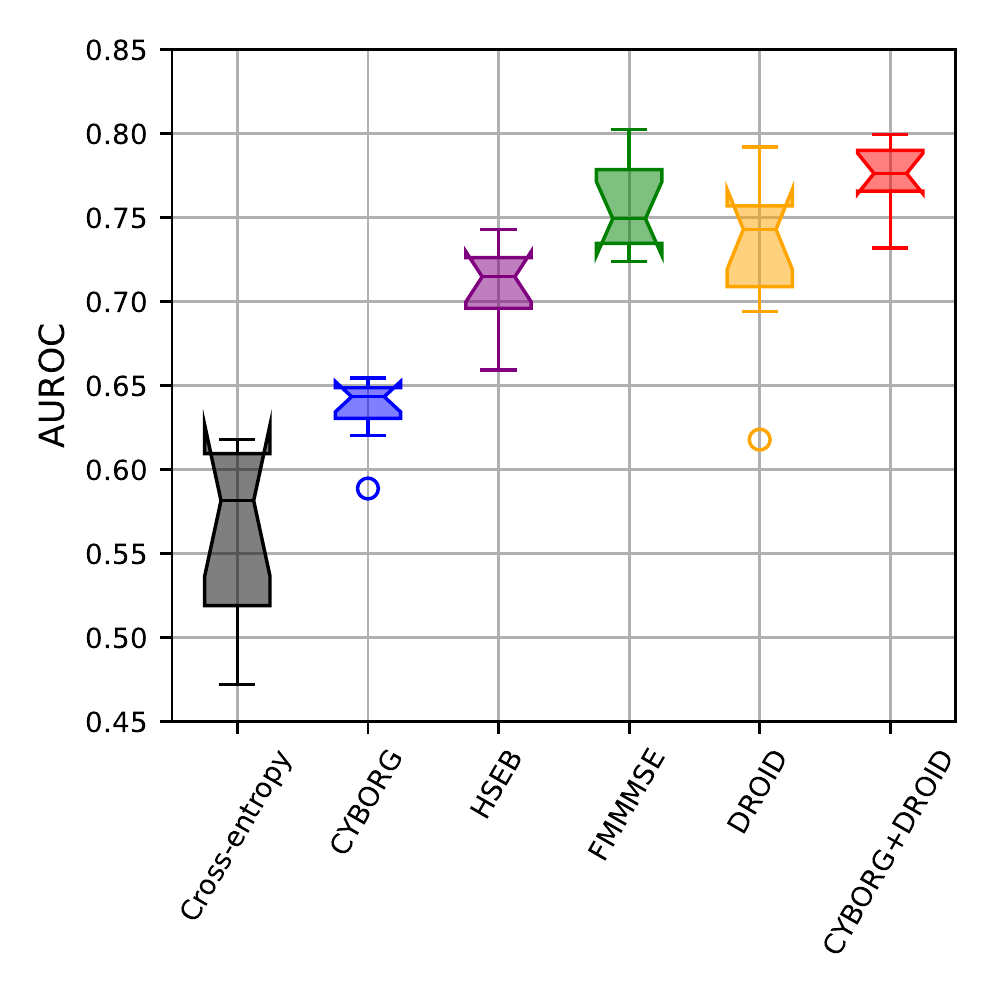}
    \caption{Boxplots representing the sigmoid AUROC scores over 10 training runs for each of the approaches considered in this paper. Thick central bars represent median values, height of each box corresponds to an inter-quartile range (IQR) spanning from the first (Q1) to the third (Q3) quartile, whiskers span from Q1-1.5$\times$IQR to Q3+1.5$\times$IQR, and outliers are shown as circles. Notches represent 95\% confidence intervals of the median. Note that for the y-axis we show only that range for which we have data, 0.45 to 0.85.
    %Note that the y-axis starts at 0.5. This is because an AUROC score near 0.5 in a binary classification indicates random chance and values lower than 0.5 indicate anti-correlation. Given the high values of our proposed methods, 0.5 serves as a more useful lower bound.
    }
    \label{fig:model_aucs}
\end{figure}

\begin{figure*}
\centering
     \begin{subfigure}{0.3\textwidth}
     \centering
         \includegraphics[width=\textwidth]{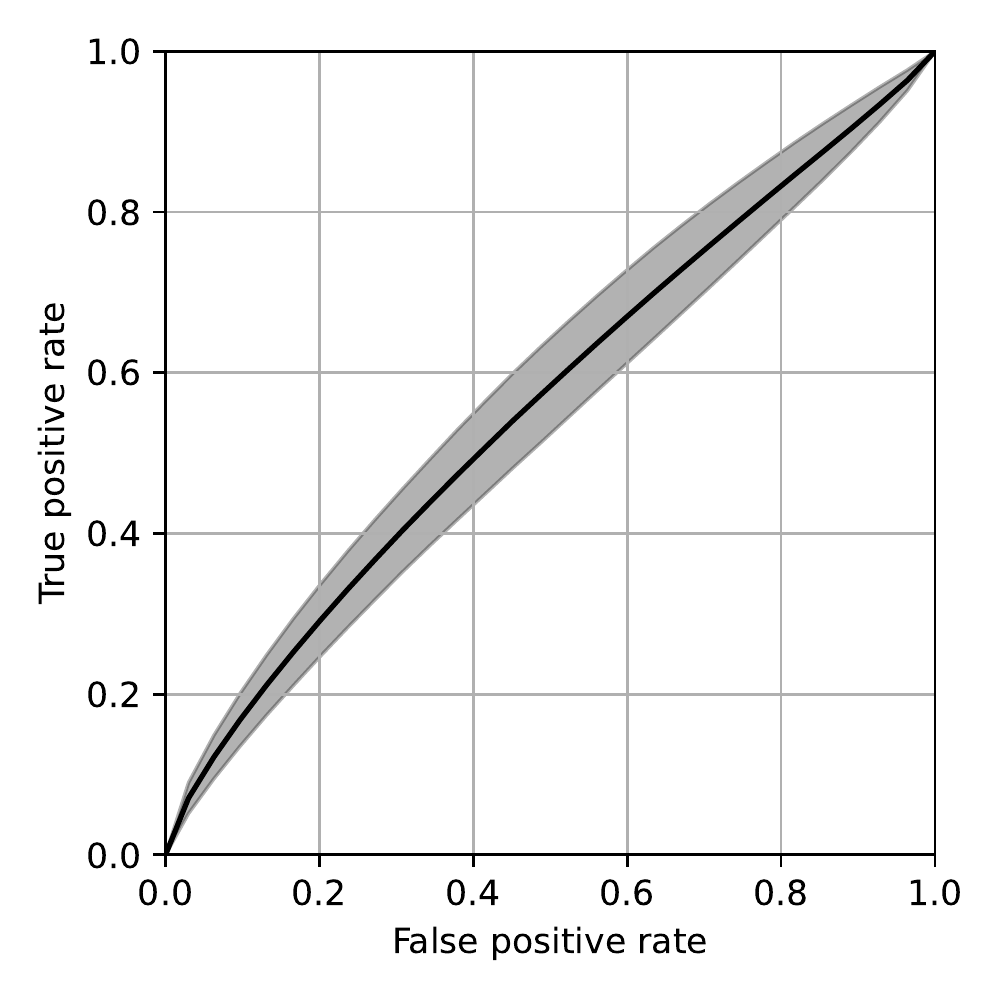}
         \caption{Cross-entropy}
         \label{fig:dist_xent}
     \end{subfigure}
     \begin{subfigure}{0.3\textwidth}
     \centering
         \includegraphics[width=\textwidth]{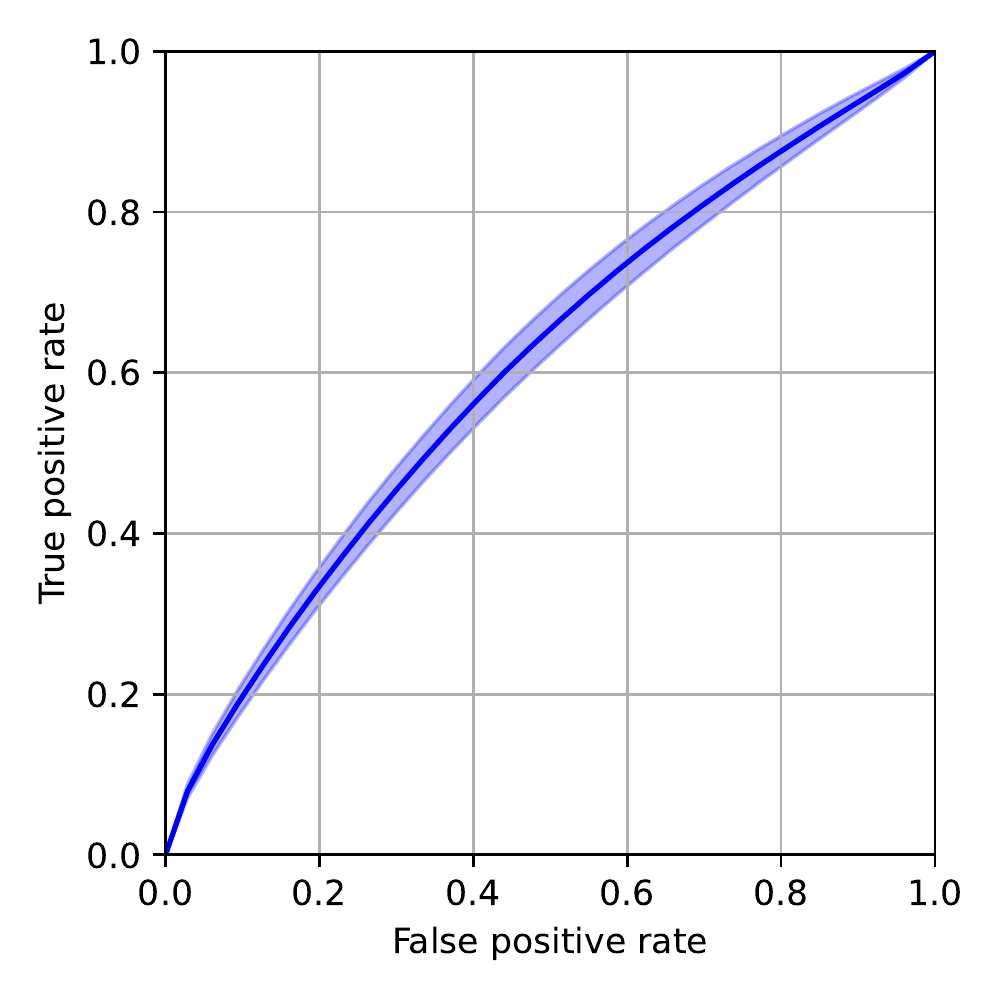}
         \caption{CYBORG}
         \label{fig:dist_cyborg}
     \end{subfigure}
     \begin{subfigure}{0.3\textwidth}
     \centering
         \includegraphics[width=\textwidth]{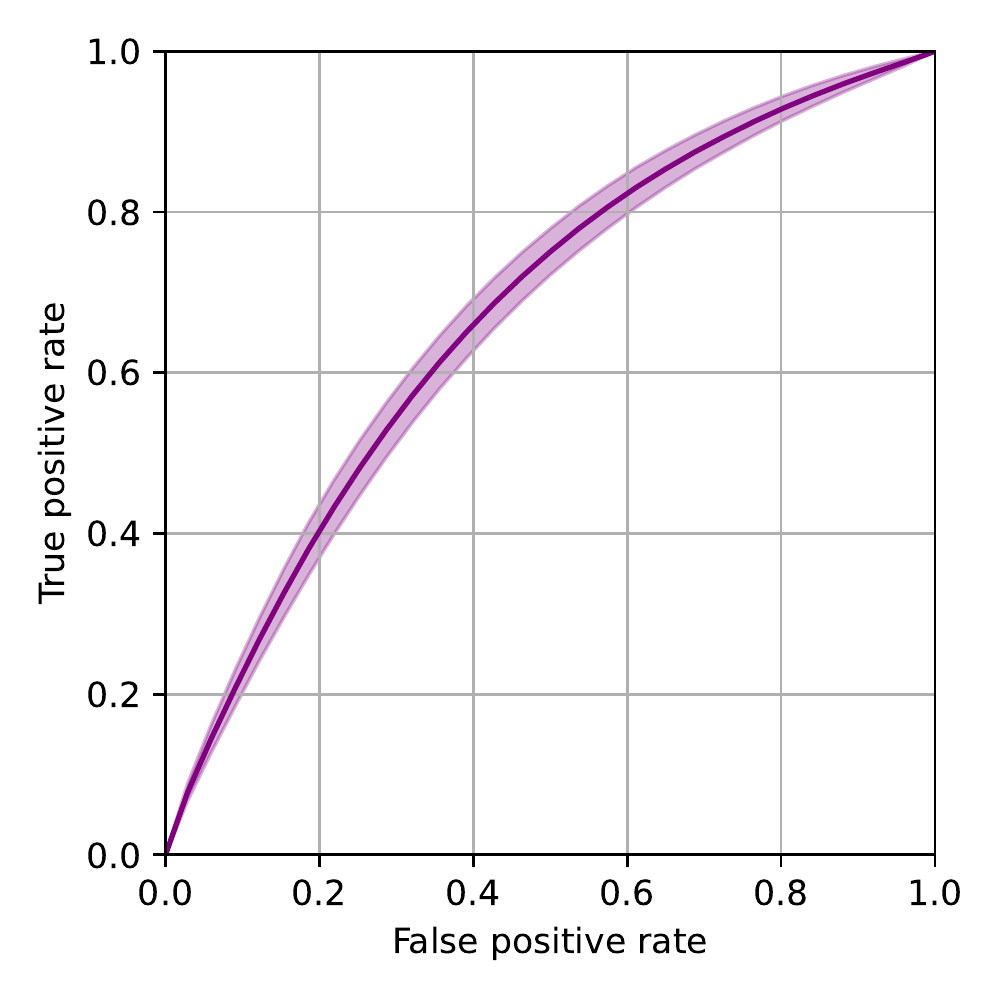}
         \caption{HSEB}
         \label{fig:dist_HSEB}
     \end{subfigure}
     
     \begin{subfigure}{0.3\textwidth}
     \centering
         \includegraphics[width=\textwidth]{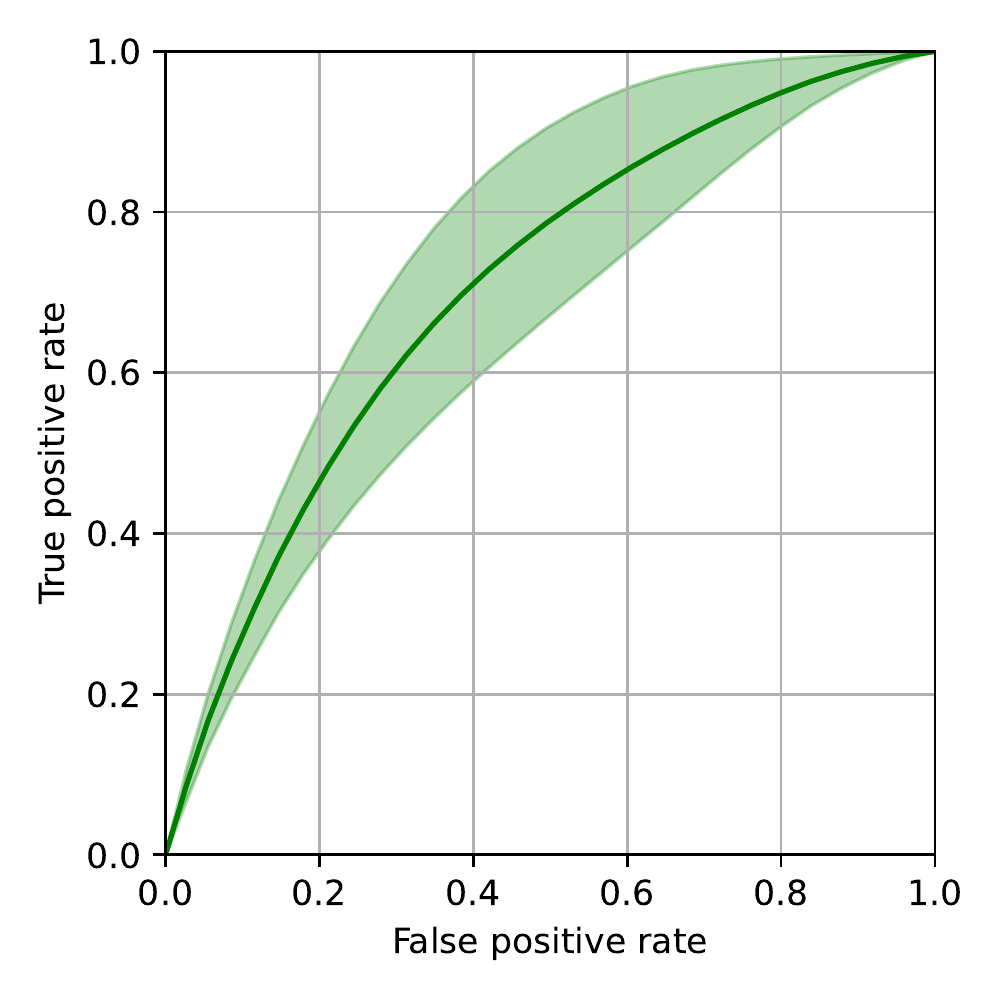}
         \caption{FMMMSE}
         \label{fig:dist_FMMMSE}
     \end{subfigure}
     \begin{subfigure}{0.3\textwidth}
     \centering
         \includegraphics[width=\textwidth]{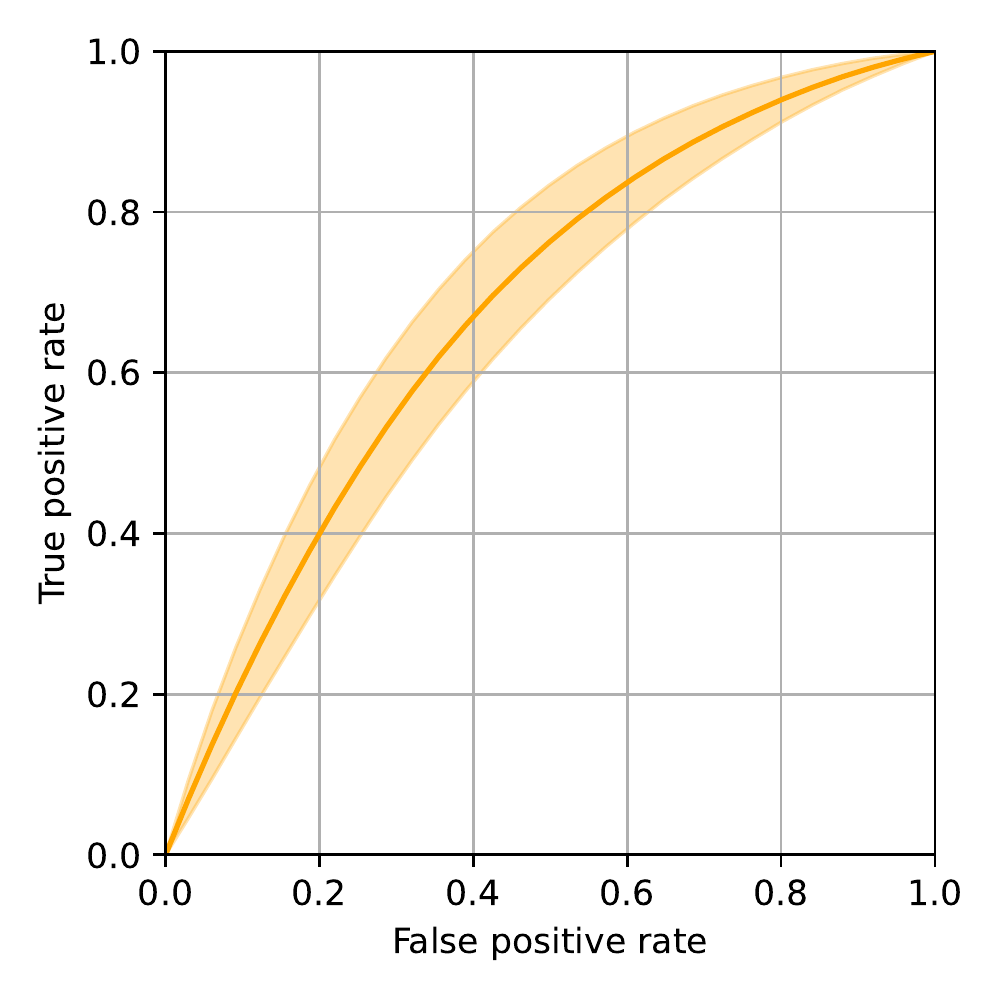}
         \caption{DROID}
         \label{fig:dist_DROID}
     \end{subfigure}
     \begin{subfigure}{0.3\textwidth}
     \centering
         \includegraphics[width=\textwidth]{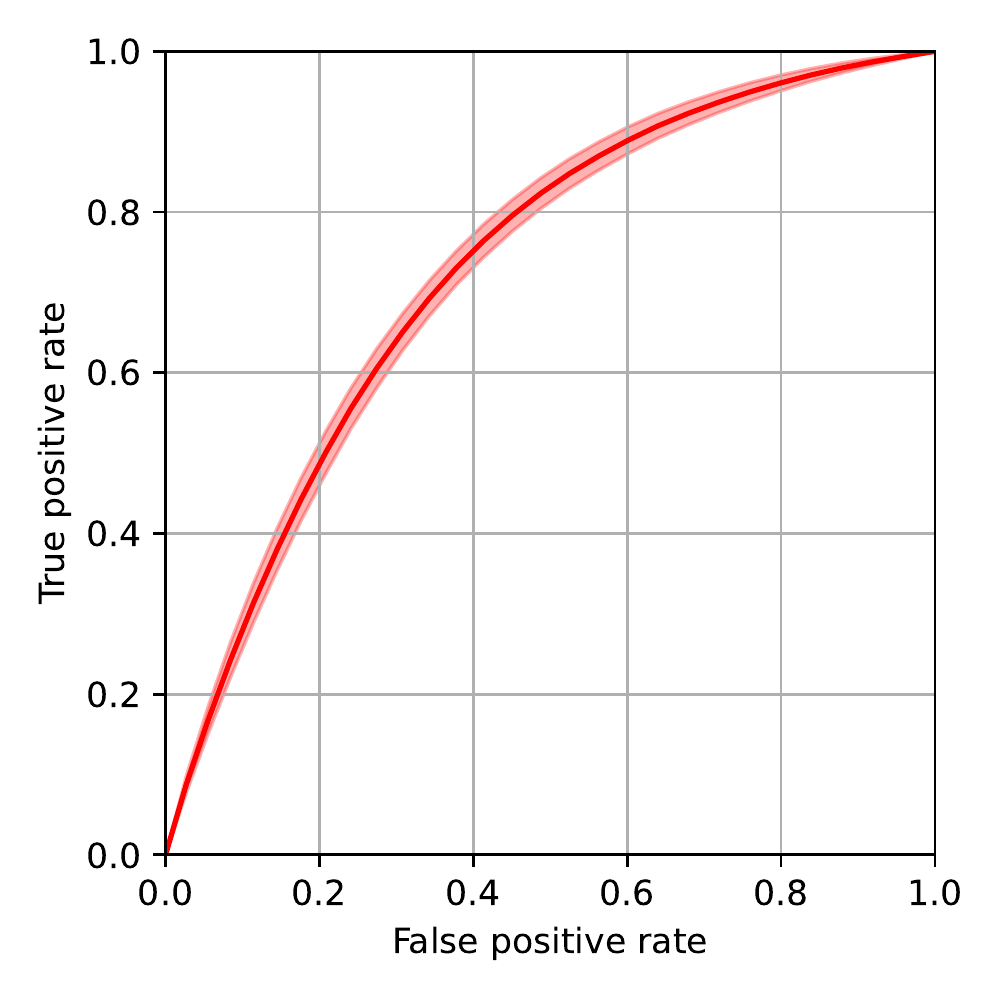}
         \caption{CYBORG+DROID}
         \label{fig:dist_cyborgdroid}
     \end{subfigure}
    \caption{Graphs showing the average Receiver Operating Characteristic curves of each of the six models tested in this paper along with the standard error of the true positive rates. It can be clearly seen that the combination of human saliency-based guidance (CYBORG) with a non-aggressive, and proposed in this paper minimization of the model salience's entropy, offers the best performance and low variability over 10 training runs.}
    \label{fig:auroc_curves}
\end{figure*}

The AUROC results supporting the answers to research questions \textbf{RQ1-4}, along with their baseline and state-of-the-art comparisons, are shown in Fig. \ref{fig:model_aucs}. Actual ROC curves are shown in Fig. \ref{fig:auroc_curves}.

\subsubsection{Answering RQ1: Does matching the model and average human salience entropies increase the performance?}
\label{sec:droid_centered_results}
In CYBORG-trained models, human saliency-guided model learns by focusing on important features. In HSEB-trained models, human saliency provides a target entropy score. Our experiments show that HSEB is able to achieve this mean entropy and it outperforms the baseline cross-entropy. Hence, \textbf{the answer to RQ1 is affirmative: requesting the model CAM entropy match human saliency entropy increases the performance}, compared to traditional cross-entropy trained models. HSEB achieves an AUROC of $0.710±0.02$. It outperforms baseline cross-entropy ($0.561±0.05$) and CYBORG ($0.636±0.02$) with a mean AUROC score increase of $+0.149$ and $+0.074$ (26.6 and 11.6 percent increase), respectively.

\subsubsection{Answering RQ2: Does unconstrained minimization of model's saliency help in achieving even better performance?}

As the answer to \textbf{RQ1} is affirmative, %we press the idea of lowering CAM entropy to its farthest extent (while still respecting validation accuracy). 
in FMMMSE-trained models, we aggressively request the minimum possible CAM entropy.
This results in the lowest mean CAM entropy models and further increases the performance, allowing us to answer the RQ2 affirmatively as well: \textbf{there well-localized and strong-enough image features that are sufficient to solve the synthetic face detection task.} The FMMMSE approach achieves an AUROC of $0.758±0.03$. It outperforms baseline cross-entropy and CYBORG with a mean AUROC score increase of $+0.197$ and $+0.122$ (35.1 and 19.2 percent increase), respectively. 

\subsubsection{Answering RQ3: Is there a better strategy to control the entropy of the model's saliency?}
As \textbf{RQ2} is affirmative, and the concern for over-focusing using the FMMMSE approach is high (due to the low percent of image focused on, and apparent high variance of the performance seen in different training runs, depicted in Fig. \ref{fig:entropy_by_model}), what if we request a middle-of-the-road CAM entropy in model training using log-entropy? 
DROID fills this middle-of-the-road position, having final mean CAM entropy scores ranging generally between FMMMSE and HSEB entropy scores. 
DROID also outperforms the baseline indicating that \textbf{a softer request of minimizing mean CAM entropy improves performance in the task of synthetic face detection.} 
DROID achieves an AUROC of $0.731±0.05$. It outperforms baseline cross-entropy and CYBORG with a mean AUROC score increase of $+0.170$ and $+0.095$ (a 30.3 and 14.9 percent increase), respectively.

\subsubsection{Answering RQ4: Does the model benefit from combining the human saliency-guided training with non-human-guided control of the model salience's entropy?}
As the results are affirmative for \textbf{RQ1-3}, and due to concerns with over-focusing in FMMMSE, we investigate the performance of a model using the combination of human saliency-guided training (CYBORG) with low entropy-based training (DROID).
The increase in performance of the CYBORG+DROID approach over baseline cross-entropy indicates that \textbf{the model benefits from combining human-guided saliency training with non-human-guided control of the model's CAM entropy.} 
CYBORG+DROID achieves our highest AUROC performance, $0.775±0.02$. It outperforms baseline cross-entropy and CYBORG by a mean AUROC score increase of $+0.216$ and $+0.141$ (a 38.5 and 22.2 percent increase) respectively.

\section{Conclusions}

High Shannon entropy of model saliency (CAM entropy) corresponds to a low focus as the model considers all pixels, including the irrelevant ones, with equal probability. 
Thus models with high entropy are indiscriminate and low-information.
This is seen with models trained with the classical cross-entropy loss function.
As CYBORG introduces human saliency to the model we can expect that the entropy decreases with the increased information.
We make the observation that this is so, leading to the natural question ``is low entropy merely an effect or can it be a cause of increased information and performance?''

This paper is an attempt to answer that question by introducing new loss functions that modify CAM entropy directly.
HSEB matches the average human-salience entropy, FMMMSE forcibly minimizes CAM entropy, and DROID seeks a reasonable middle ground by minimizing log entropy.
Indeed we see AUROC improvements in all three methods as we reduce CAM entropy.

The next question raised is ``how far can we reduce CAM entropy before the information gain becomes a hindrance to the model?'' 
In Fig. \ref{fig:entropy_by_model} we see that no model achieves its highest AUROC performance with a CAM entropy below 1.0.
Instead, the best performances are generally in the entropy range of 2.0-2.5.
This leads us to consider the middle-of-the-road DROID method as the most optimal.

Finally, as the incorperation of human-salience has proven useful in the past, it stands to reason that human direction could help guide the more focused, low-entropy DROID method.
This leads us to the final question of this paper, ``does incorporating human-salience into an optimal low-entropy model improve performance?''

The answer is yes.
CYBORG+DROID acheives the highest average AUROC of all the methods in this paper, improving over DROID by $+0.024$.
While this difference appears significant (Fig. \ref{fig:model_aucs}), it is ultimately quite small (a $0.03$ percent increase) indicating that there is a need for further work in combining model-salience entropy and human-salience.

%Human saliency-guided CYBORG indirectly reduces the Shannon entropy of model saliency over the course of it's training. Our proposed low-entropy models directly reduce CAM entropy and see an increase in sigmoid AUROC performance over baseline cross-entropy and state-of-the-art CYBORG. 
%However, we indicate that there is a risk of a model over-focusing when it has too low of an entropy, but this only appears in FMMMSE models which aggressively reduce entropy to its lowest value (while maintaining accuracy). 
%HSEB, which matches human entropy values, and DROID, which moderately reduces entropy using a log entropy loss function, increase AUROC performance while not over-focusing. 
%Of our three low entropy models DROID performs better, thus we attempt to combine it with human saliency-guided training. The end result, CYBORG+DROID performs best of all achieving an AUROC score 0.775±0.02 against baseline 0.561±0.05. There is potential for low entropy models in the task of synthetic face detection and even more in the combination of low entropy and human saliency-guided methods. However, future work must be done to explore these methods in other task domains.
%CYBORG reduces cam entropy and that's a good thing. FMMMSE demonstrates that it is a good thing by reducing the cam entropy even farther and improving performance. However, it goes too far and leads to over training. Finally, DROID seems to strike a good balance. Maybe we talk about combining CYBORG and DROID to see if the directing-aspect of CYBORG is an improvement over DROID, or maybe we leave it for future work.

\printbibliography

\end{document}